\documentclass{article}
\usepackage{spconf,amsmath,graphicx}
\usepackage{times}
\usepackage{soul}
\usepackage{url}
\usepackage[hidelinks]{hyperref}
\usepackage[utf8]{inputenc}
\usepackage[small]{caption}
\usepackage{graphicx}
\usepackage{amsmath}
\usepackage{amsthm}
\usepackage{booktabs}
\usepackage{algorithm}
\usepackage{enumerate}
\usepackage{bbm}
\usepackage{diagbox}
\usepackage{multirow}
\usepackage{arydshln}
\usepackage[noend]{algpseudocode}
\usepackage{amssymb}
\usepackage[switch]{lineno}
\usepackage[dvipsnames]{xcolor}
\usepackage{algorithmicx,algorithm}

\usepackage{color}

\newcommand{\setParDis}{\setlength {\parskip} {1cm} }
\newcommand{\setParDef}{\setlength {\parskip} {0pt} }

\title{TEAM DETR: GUIDE QUERIES AS A PROFESSIONAL TEAM \\IN DETECTION TRANSFORMERS}
\name{Tian Qiu$^{\dagger}$, Linyun Zhou$^{\dagger}$, Wenxiang Xu$^{\dagger}$, Lechao Cheng$^{\dagger\dagger}$\sthanks{Corresponding author.}, Zunlei Feng$^{\dagger}$, Mingli Song$^{\dagger}$}
\address{$^{\dagger}$ Zhejiang University\\ $^{\dagger\dagger }$ Zhejiang Lab}
\begin{document}
\ninept
\maketitle
\begin{abstract}
Recent proposed DETR variants have made tremendous progress in various scenarios due to their streamlined processes and remarkable performance. However, the learned queries usually explore the global context to generate the final set prediction, resulting in redundant burdens and unfaithful results. More specifically, a query is commonly responsible for objects of different scales and positions, which is a challenge for the query itself, and will cause spatial resource competition among queries. To alleviate this issue, we propose Team DETR, which leverages query collaboration and position constraints to embrace objects of interest more precisely. We also dynamically cater to each query member's prediction preference, offering the query better scale and spatial priors. In addition, the proposed Team DETR is flexible enough to be adapted to other existing DETR variants without increasing parameters and calculations. Extensive experiments on the COCO dataset showcase that Team DETR achieves remarkable gains, especially for small and large objects. Code is available at \url{https://github.com/horrible-dong/TeamDETR}.

\end{abstract}
\section{Introduction}
Object detection is an essential task in computer vision, widely used in face recognition, autonomous driving, and security monitoring \cite{detsurvey1,detsurvey2,detsurvey3,detsurvey4}. Most classical detectors, such as the R-CNN \cite{rcnn,fastrcnn,fasterrcnn,cascadercnn} and YOLO \cite{yolov1,yolov2,yolov3,yolov4} series, are of convolutional architectures and have achieved remarkable performance. However, these detectors have disadvantages such as complex structure, dense prediction, and non-end-to-end. Recently, DETR (DEtection TRansformer) \cite{detr} has introduced a more straightforward end-to-end approach, modeling object detection as a set prediction problem without NMS.

Many follow-up works are devoted to increasing query interpretability. A query is now decoupled into a content and a spatial one \cite{conditionaldetr}. The spatial query is modeled as an anchor point \cite{conditionaldetr,anchordetr} or anchor box \cite{dabdetr}, continuously updated during training. However, the current works only give a query an explicit physical meaning. The division of labor for queries is still unclear. A query is commonly responsible for objects of different scales and positions. It is a challenge for the query itself, and there will also be spatial resource competition among queries.

In this paper, we are dedicated to effectively guiding the queries as a professional team. Without increasing parameters and calculations, we strengthen the division of labor among queries by assigning functions to queries in terms of scale and space.

\begin{figure}[!t]
    \centering
    \includegraphics[scale =0.44]{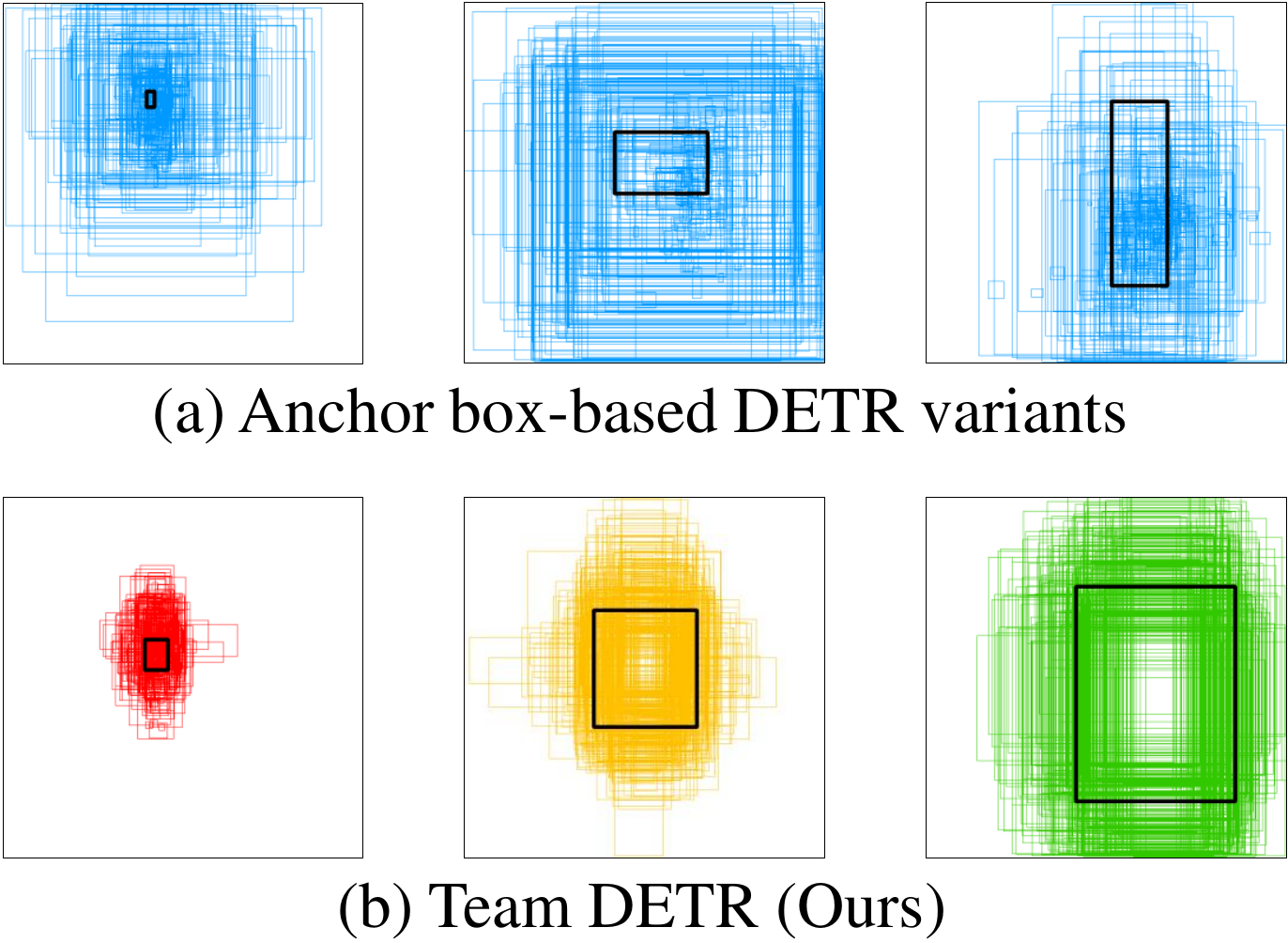}
    \caption{
    The prediction box distribution of queries with different anchor box scales. In anchor box-based DETR variants (a) \cite{dabdetr,dndetr,dino}, a spatial query is specified as an anchor box (in the figure, the black box). Without particular constraints on query behavior, it will result in excessive prediction scale variance. Besides, the management area of each query is relatively large. When two anchor boxes are close together, it will inevitably cause competition for spatial resources. Our proposed Team DETR (b) addresses the above issues by specifying the division of labor for queries.
    }
    \label{fig:fig1}
\end{figure}

As shown in Fig.\ref{fig:fig1}(a), the same query will be assigned to objects with excessive scale variance in different images, which poses a challenge for the learning process. We group queries at the decoder side and specify that each group is only responsible for objects within a specific scale range. The queries that are responsible for the same scale range will work together to find the optimal match. Moreover, also shown in Fig.\ref{fig:fig1}(a), queries are all in charge of a relatively large area. It leads to resource competition at the edges of their assigned areas, which hinders the team's development. For this reason, we present position constraints to make their respective areas of responsibility more focused and avoid unnecessary internal conflicts. Furthermore, we dynamically extract each query's prediction preferences, providing it with improved scale and spatial priors.


Our contribution is the proposal of a Team DETR, which incorporates a new framework for query collaboration, guiding queries as a professional team. As a necessary complement to query interpretability, in terms of scale and spatial position, the functions of team members are assigned and constrained in a reasonable manner and are dynamically adjusted to accommodate the members' preferences. The proposed Team DETR can be seamlessly integrated into other existing DETR variants without any increase in parameters or calculations and achieves significant improvements on the COCO \cite{coco} benchmark, particularly for small and large objects, which demonstrates its effectiveness and generalization. Fig. \ref{fig:fig1}(b) shows the prediction box distribution of queries in Team DETR.

\begin{figure*}[!t]
    \centering
    \includegraphics[scale =0.44]{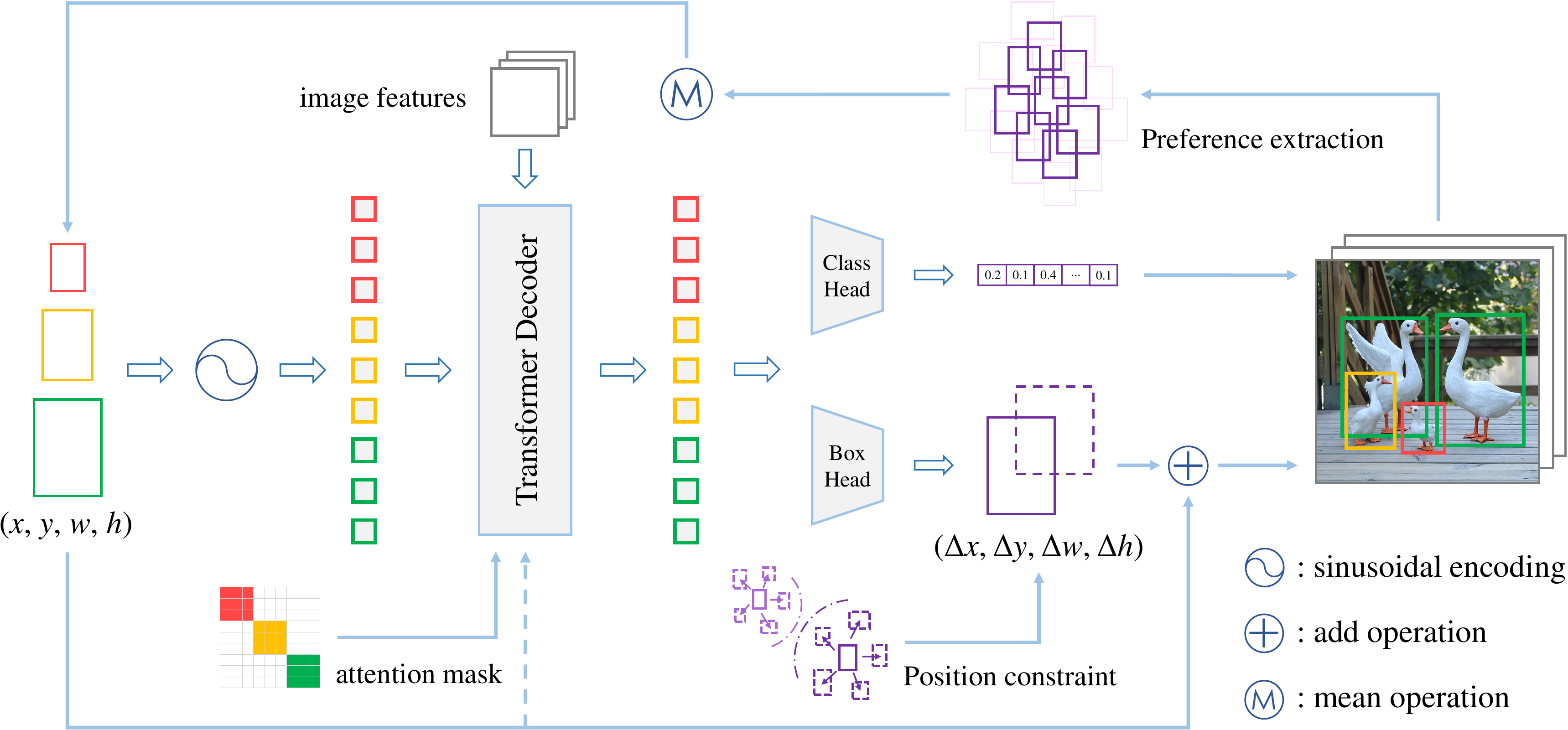}
    \caption{
    The framework of the proposed Team DETR, which is based on the basic architecture of DAB-DETR. The CNN backbone is used to extract image features, which are then fused by the transformer encoder. The decoder utilizes several learned queries to match objects for the image features. A query is represented as an anchor box $(x, y, w, h)$ and is dynamically updated based on the offset $(\Delta x, \Delta y, \Delta w, \Delta h)$ predicted by each layer of the decoder. Building upon this, we introduce a query teamwork approach in which the queries are grouped, and each group is responsible for objects within a specific scale range. To avoid resource competition, the management area of each query is limited. Furthermore, the prediction preferences of each query are dynamically extracted, and the anchor is updated accordingly.
    }
    \label{fig:fig2_framework}
\end{figure*}

\section{Related Work}
\textbf{Classical anchor-based detectors}.
\setParDis
Classical object detectors that use anchor-based architecture have a spatial prior reinforced by the preset anchors. In the past, the term ``anchor" specifically referred to the anchor box, and the methods that do not use anchor boxes were collectively called the anchor-free method. However, as technology evolves, the definition of the term ``anchor" has expanded to encompass both anchor boxes and anchor points. The anchor box-based classical detectors include the R-CNN series \cite{rcnn,fastrcnn,fasterrcnn,cascadercnn}, SSD \cite{ssd}, YOLOv2-v4 \cite{yolov2,yolov3,yolov4}, and RetinaNet \cite{retinanet}. These models use dense anchor boxes of varying scales to accommodate objects of varying sizes and positions. Detectors such as YOLOv1 \cite{yolov1}, FCOS \cite{fcos}, FoveaBox \cite{foveabox}, and others \cite{centernet,tridentnet,softanchorpoint} start from the 2D anchor point and directly predict the distance from the point to the box's border. Due to the massive spatial priors, anchor-based classical detectors significantly improve the convergence speed and accuracy.

\setlength{\parskip}{1em}
\noindent\textbf{DETR series}.
Recently, DETR \cite{detr} has provided a new solution for object detection. The DETR encoder fuses image features, and the decoder matches the objects from the image features by defining 100 queries. The one-to-one assignment does not need NMS and achieves the true end-to-end. However, the original DETR converges very slowly. Subsequent improvements to DETR mainly focus on the query's interpretability. Conditional DETR \cite{conditionaldetr} decouples the query into a content and a spatial one, which significantly accelerates the convergence of the model. In Conditional DETR \cite{conditionaldetr} and Anchor DETR \cite{anchordetr}, the spatial query is obtained by mapping an anchor point to a high-dimensional space, which provides a better position prior. DAB-DETR \cite{dabdetr} offers a deeper understanding of the role of queries. The 2D anchor point is extended to the 4D anchor box and dynamically updates layer by layer in the decoder. Based on DAB-DETR, DN-DETR \cite{dndetr} introduces query denoising to accelerate convergence. DINO \cite{dino} further introduces contrastive denoising to achieve state-of-the-art. The above methods have contributed a lot to the interpretability of the query itself. Nevertheless, the division of labor for queries is still unclear. Our work is to address this issue.
\setParDef

\section{Method}
We propose Team DETR, which improves queries on the decoder without adding extra parameters or calculations. The core idea is to function queries as a team, making team management crucial. The division of labor among query members needs to be clarified. Considering the object scale, we group query members, and each group is responsible for predicting objects within a scale range. The group members work together to determine the optimal match for the target objects. In terms of spatial position, we impose positional constraints on the team members to limit their focus to specific areas in the image. To maximize the capabilities of each team member, we dynamically analyze each query's prediction preference and update the anchor accordingly. 

Fig. \ref{fig:fig2_framework} shows Team DETR’s framework. We follow the basic architecture of DETR and make improvements based on DAB-DETR. Since we only improve the spatial query, in the following, the term ``query" specifically refers to the ``spatial query". Next, we will elaborate on each part of the teamwork.

\subsection{Scale-wise grouping}
\label{sec:scale_wise_grouping}
As illustrated in Fig. \ref{fig:fig1}(a), in the absence of guidance, a query has to predict objects with excessive scale variance, which invisibly increases its learning difficulty. Therefore, we propose a scale-wise grouping approach. The original queries are divided into groups, each responsible for predicting objects within a specific scale range. 

Firstly, define $K$ scale ranges $\left\{S_k\right\}^K_{k=1}$, $S_k=\left(s_k^{min}, s_k^{max}\right].$
\noindent Here, we employ relative scales instead of absolute scales commonly used in convolutional detectors. The explanation is that the attention \cite{attention} is different from the convolution. The convolution extracts local features, while the attention extracts global features. It indicates that in transformers, using absolute scales is unreasonable because the object scale is referenced to the global context. Therefore, the relative scales are adopted. Then, the scales are normalized between 0 and 1, and the $K$ scale ranges $\left\{S_k\right\}^{K}_{k=1}$ will cover the normalized range $\left(0,1\right]$ as follows:
\begin{equation}
\bigcup_{k=1}^{K}S_k=\left(0,1\right].
\end{equation}

Correspondingly, divide queries into $K$ groups $\left\{Q_k\right\}^{K}_{k=1}$. The $k$-th group $Q_k$ is composed of $n_k$ queries $\left\{q_i\right\}^{n_k}_{i=1}$. Then, the grouped query sets $\mathbb{Q}$ can be denoted as follows:
\begin{equation}
\mathbb{Q}=\left\{Q_k\right\}^{K}_{k=1},\,\,\,\,\,\,Q_k=\left\{q_i\right\}^{n_k}_{i=1},\,\,\,\,\,\,\sum_{k=1}^{K}n_k=N,
\end{equation}
\noindent where $N$ is the total number of the original queries. In the transformer, the $i$-th query $q_i$ is obtained by mapping the anchor box $A_i$ to a high-dimensional space through sinusoidal positional encoding $PE_{sin}(\cdot)$ and multi-layer perceptron $MLP(\cdot)$ as follows:
\begin{equation}
q_i=MLP\left(PE_{sin}\left(A_i\right)\right), q_i\in\mathbb{R}^D,
\end{equation}
\noindent where the anchor box $A_i=(x_i,y_i,w_i,h_i)$. $x_i,y_i,w_i,h_i\in\mathbb{R}$, and $D$ denotes the dimension of $q_i$. When initializing anchor boxes, all anchors' center points $(x_, y)$ are randomly and uniformly distributed in the image. For the $k$-th group, the initial values of $w$ and $h$ are set to $\left(s_k^{min}+s_k^{max}\right)/\,2$.

With the scale ranges and query groups defined, we stipulate that the $k$-th group of queries $Q_k$ is responsible for ($\unrhd$) the objects within the $k$-th scale range $S_k$:
\begin{equation}
Q_k\unrhd Obj_{\in S_k}.
\end{equation}
Decomposing a problem into multiple sub-problems is the idea of divide and conquer. This way, each query’s prediction scale variance will be greatly reduced. 

After the above preparations, the queries will be fed into the transformer decoder, performing self-attention and cross-attention with image features. Self-attention allows queries to communicate with each other, preparing for the subsequent matching objects in cross-attention. Queries within the same group can communicate freely, but in order to prevent interference from irrelevant information, communication between members from different groups is restricted. Thus, we define the attention mask $M=\left[m_{ij}\right]_{N\times N}$.
\noindent For the $k'$-th group of queries, the corresponding mask element $m_{ij}$ is calculated as follows: 
\begin{equation}
m_{ij}=\left\{\begin{matrix}
\begin{aligned}
\,\,& False,        \,\,\, if \,\,\,{\textstyle \sum_{k=1}^{k'-1}n_k}<i,j\le{\textstyle \sum_{k=1}^{k'}n_k}
 \\
\,\,& True   \,\,,  \,\,\, otherwise
\end{aligned}
\end{matrix}\right..
\end{equation}
\noindent Here, ${\textstyle \sum_{k=1}^{0}n_k=0}$. $m_{ij}=True$ means the $i$-th query cannot see the $j$-th query. 

In the proposed Team DETR, with the input image $I$, the final matching results $\delta\left(\mathbb{Q},Obj\right)$ between the grouped query sets $\mathbb{Q}$ and all the objects $Obj$ is calculated as follows:
\begin{equation}
\delta\left(\mathbb{Q},Obj\right)=\biguplus_{k=1}^{K}\mathcal{H}\left(\mathcal{F}\left(Q_k,M,I\right),Obj_{\in S_k}\right),
\end{equation}
\noindent where $\mathcal{F}(\cdot, \cdot, \cdot)$ denotes the mapping function of DETR,
$\mathcal{H}(\cdot, \cdot)$ is the Hungarian matching algorithm, and $\biguplus$ means joining the matching results of each group.

\begin{table*}[!t]
  \centering
  \renewcommand\arraystretch{0.92}
  \caption{Main results for Team DETR on COCO val2017. R50 / R101 denote using ResNet-50 / ResNet-101 \cite{resnet} as the backbone. Our Team DETR is adapted to DAB-based DETRs, including DAB-DETR and its follow-ups, DN-DETR, and the current state-of-the-art DINO. They all achieve impressive boosts, especially for small and large objects.}
  \resizebox{1\linewidth}{!}{
    \begin{tabular}{lcccccccccc}
    \toprule
    Model & w/ Team DETR & Epochs & AP    & AP$_{50}$ & AP$_{75}$  & AP$_s$   & AP$_m$   & AP$_l$   & Params & GFLOPs \\
    \midrule
    DETR-R50 \cite{detr} &       & 12    & 21.1  & 37.9  & 20.4  & 6.3   & 21.4  & 34.9  & 41M   & 86 \\
    Anchor DETR-R50 \cite{anchordetr} &       & 12    & 30.8  & 51.3  & 31.9  & 13.9  & 33.9  & 45.2  & 37M   & - \\
    Conditional DETR-R50 \cite{conditionaldetr} &       & 12    & 32.2  & 51.7  & 33.6  & 14.1  & 34.6  & 48.2  & 44M   & 90 \\
    \midrule
    DN-DETR-R50 \cite{dndetr} &       & 12    & 37.3  & 57.5  & 39.2  & 17.2  & 40.1  & 55.6  & 44M   & 94 \\
    DN-DETR-R50 \cite{dndetr} & \checkmark     & 12    & \textbf{37.7(+0.4)} & 57.8  & 39.6  & \textbf{18.0(+0.8)} & 40.0  & \textbf{56.8(+1.2)} & 44M   & 94 \\
    \midrule
    DAB-DETR-R50 \cite{dabdetr} &       & 12    & 33.7  & 54.1  & 35.1  & 15.3  & 36.5  & 49.7  & 44M   & 94 \\
    DAB-DETR-R50 \cite{dabdetr} & \checkmark     & 12    & \textbf{35.3(+1.6)} & 56.3  & 36.5  & \textbf{17.3(+2.0)} & 37.5  & \textbf{52.9(+3.2)} & 44M   & 94 \\
    \midrule
    DAB-DETR-R50 \cite{dabdetr} &       & 50    & 42.2  & 62.8  & 44.8  & 22.5  & 45.9  & 60.2  & 44M   & 94 \\
    DAB-DETR-R50 \cite{dabdetr} & \checkmark     & 50    & \textbf{42.9(+0.7)} & 63.9  & 45.6  & \textbf{24.1(+1.6)} & 46.0  & \textbf{62.4(+2.2)} & 44M   & 94 \\
    \midrule
    DAB-DETR-R101 \cite{dabdetr} &       & 12    & 36.1  & 56.5  & 38.2  & 17.3  & 39.5  & 52.5  & 63M   & 174 \\
    DAB-DETR-R101 \cite{dabdetr} & \checkmark     & 12    & \textbf{37.4(+1.3)} & 58.4  & 39.4  & \textbf{18.4(+1.1)} & 40.3  & \textbf{55.5(+3.0)} & 63M   & 174 \\
    \midrule
    DAB-DETR-R101 \cite{dabdetr} &       & 50    & 43.3  & 64.0  & 46.7  & 24.0  & 47.1  & 61.2  & 63M   & 174 \\
    DAB-DETR-R101 \cite{dabdetr} & \checkmark     & 50    & \textbf{44.1(+0.8)} & 65.1  & 47.0  & \textbf{25.0(+1.0)} & 47.1  & \textbf{63.7(+2.5)} & 63M   & 174 \\
    \midrule
    DINO-4scale-1stage-R50 \cite{dino} &       & 12    & 44.5  & 61.7  & 48.2  & 24.2  & 48.0  & \textbf{61.2 } & 47M   & 236 \\
    DINO-4scale-1stage-R50 \cite{dino} & \checkmark     & 12    & \textbf{46.3(+1.8)} & 63.5  & 50.6  & \textbf{28.6(+4.4)} & 48.9  & \textbf{61.2(+0.0)} & 47M   & 236 \\
    \bottomrule
    \end{tabular}
    }
  \label{tab:tab1}%
\end{table*}%

\subsection{Position constraint}
\begin{figure}[!t]
    \centering
    \includegraphics[scale =0.42]{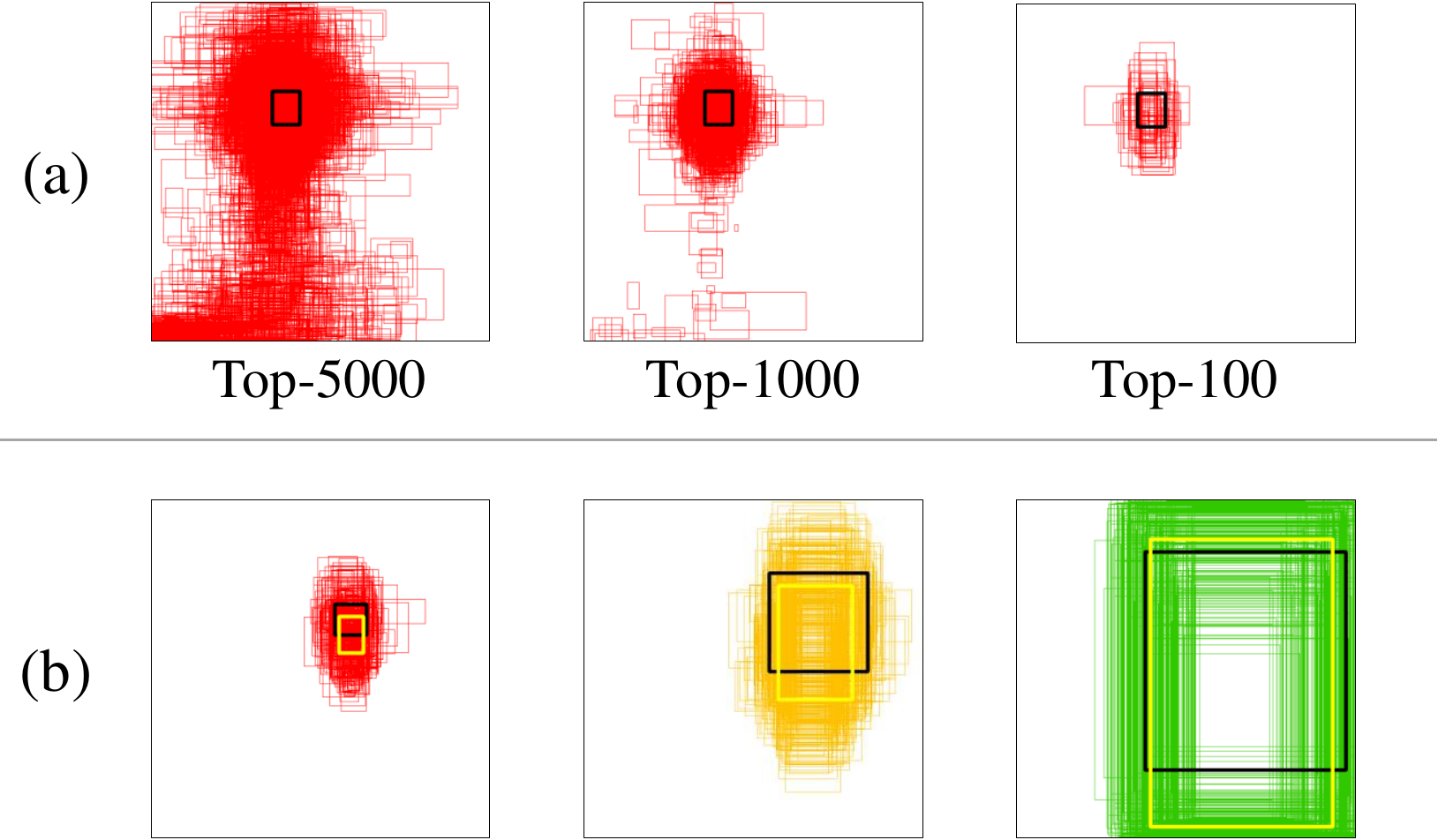}
    \caption{(a) Box predictions of a query. The prediction boxes with higher confidence rankings tend to be distributed near the center of the anchor box. (b) Each query has its preference for the prediction box's scale, shape, and position. The black box is the initial anchor, and the yellow box is the updated anchor obtained by taking the mean value of the prediction boxes with high confidence, representing the query's prediction preference.
    }
    \label{fig:fig3}
\end{figure}

Initializing the query with the anchor box can obtain the spatial position prior. As shown in Fig. \ref{fig:fig3}(a), we observe that the prediction boxes with higher confidence rankings tend to be distributed near the center of the anchor box. When queries are close to each other, resource competition is likely to occur at the edge of the management area due to the high number of queries. These internal conflicts result in low confidence rankings of prediction boxes near the edges. Therefore, we limit the predictions of the queries to a specific area close to the center of the anchor box to reduce competition and increase the number of prediction boxes with higher confidence rankings.

Let $\hat{B}_i$ be the prediction box of the $i$-th query $q_i$. When the distance between the center points of the prediction box $\hat{B}_i$ and the anchor box $A_i$ exceeds the threshold $\eta$, a penalty is imposed on $\hat{B}_i$. Then, the position loss $\mathcal{L}_{pos}$ is expressed as follows:
\begin{equation}\footnotesize
\begin{split}
\mathcal{L}_{pos}=\frac{1}{\sigma}\sum_{i=1}^{N}\mathbbm{1}_{\left\{\left\|\hat{B}_i^{\left\{x,y\right\}}-A_i^{\left\{x,y\right\}}\right\|_2>\eta\right\}}\left\|\hat{B}_i^{\left\{x,y\right\}}-A_i^{\left\{x,y\right\}}\right\|_2,
\end{split}
\end{equation}
\noindent where $\sigma$ is the number of boxes to be penalized. 

\subsection{Preference extraction}
Team members will bring out their respective strengths when performing tasks. As shown in Fig. \ref{fig:fig3}(b), each query has its preference for the prediction box's scale, shape, and position. In order to better tap their potential, we extract the query's prediction preference from the prediction boxes with high confidence and use this preference as the new anchor box. Preference extraction dynamically updates the prior. It is performed in the validation process after each epoch. 

Let $\hat{\mathbb{B}}_i$ be the set of prediction boxes of $q_i$ in all validation images. Pick out the $\tau$ boxes with the highest confidence from $\hat{\mathbb{B}}_i$, and take their mean value to get the fused box $\check{A}_i$.$\check{A}_i$ is $q_i$'s preference, used as the updated anchor box $A_i$. It can be expressed as
\begin{equation}
A_i\leftarrow\check{A}_i=\frac{1}{\tau}\sum\hat{\mathbb{B}}_i^{\downarrow_{\tau_{conf}}}.
\end{equation}
\noindent This statistical process does not introduce the validation label and is done by the way during the validation, whose calculation cost is negligible. In summary, preference extraction is designed to cater to the query's personality, which ultimately benefits the query itself.

\subsection{Training and optimization}
The above query teamwork can be easily integrated into other existing DETR variants. As a new loss item, $\mathcal{L}_{pos}$ is summed with the class 
 loss $\mathcal{L}_{cls}$ and the box losses, $\mathcal{L}_{l1}$ and $\mathcal{L}_{giou}$. The final total loss $\mathcal{L}$ is expressed as follows:
\begin{equation}
\mathcal{L}=\lambda_{cls}\mathcal{L}_{cls}+\lambda_{l1}\mathcal{L}_{l1}+\lambda_{giou}\mathcal{L}_{giou}+\lambda_{pos}\mathcal{L}_{pos},
\end{equation}
where $\lambda_{\{cls, l1, giou, pos\}}$ are the balanced parameters.

\section{Experiments}
\subsection{Dataset and experiment setup}
The challenging COCO 2017 dataset \cite{coco} is used to validate our method. We divide the queries into three groups, with the proportion of 65\%, 20\%, and 15\%, corresponding to the relative scales of (0, 0.2], (0.2, 0.4], and (0.4, 1], respectively. The values of $\eta$, $\lambda_{pos}$, $\tau$ are set to 0.25, 5, and 300, respectively. The learning rate decreases by multiplying 0.1 at the 40-th epoch for the 50-epoch setting and the 8-th epoch for the 12-epoch setting. All other settings in each set of comparative experiments are kept identical.
 
\subsection{Detection result comparison}
Table~\ref{tab:tab1} shows the detection result compared with some SOTA methods on COCO val2017. Our query teamwork method adapts well to DAB-based DETRs, including DAB-DETR \cite{dabdetr} and its successors DN-DETR \cite{dndetr}, and the current state-of-the-art DINO \cite{dino}. The proposed method achieves impressive improvement, especially for small and large objects. Under the 12-epoch setting (also known as the 1x setting), our method gains +1.6 AP / +1.3 AP on DAB-DETR-R50 and DAB-DETR-R101, among which +2.0 AP / +1.1 AP for small objects and +3.2 AP / +3.0 AP for large objects. Under the 50-epoch setting, our method on DAB-DETR boosts more than 1 AP for small objects and more than 2 AP for large objects. DINO is currently the strongest DETR-based detector, and our Team DETR improves the single-stage DINO's performance by up to 1.8 AP (an improvement of +4.4 AP for small objects). What's more the proposed method does not increase parameters and calculations. The experimental results effectively verify the effectiveness and generalization of the proposed Team DETR.

\subsection{Ablation study}
\begin{table}[!t]\small
  \centering
  \renewcommand\arraystretch{1}
  \caption{Ablation study on different components of Team DETR.}
    \begin{tabular}{lc}
    \toprule
    Setting & AP \\
    \midrule
    S1: \,\,\,DAB-DETR-R50 & 33.7 \\
    S2: \,\,\,S1 + Scale-wise grouping (absolute scales) & 34.4 \\
    S3: \,\,\,S1 + Scale-wise grouping (relative scales) & 34.9 \\
    S4: \,\,\,S3 + Position constraint & 35.1 \\
    S5: \,\,\,S4 + Preference extraction (S1 w/ Team DETR) & \textbf{35.3} \\
    \bottomrule
    \end{tabular}%
  \label{tab:tab2}%
  \vspace{-0.6cm}
\end{table}%
In this section, we conduct an ablation study on the scale-wise grouping (absolute scales), scale-wise grouping (relative scales), position constraint, and preference extraction. DAB-DETR-R50 under the 1x setting is used as the baseline. From Table \ref{tab:tab2}, we can see that scale-wise grouping increases the baseline’s accuracy from 33.7 AP to 34.9 AP. Position constraint and Preference extraction further improve the performance to the 35.3 AP. Notably, when queries are grouped based on the absolute scales rather than relative scales, there is a significant drop in performance, which supports our assertion in Section~\ref{sec:scale_wise_grouping} that the relative scale is more appropriate for transformers that extract global features.

\section{Conclusion}
In this paper, we propose Team DETR, which clarifies the division of labor for queries, as a way to enhance the query's interpretability. We view queries as a professional team and elaborately assign roles to them according to their scale and spatial responsibilities. The proposed Team DETR addresses two major issues that arise from the unclear division of labor among queries: excessive variance in prediction scales and competition for spatial resources. Team DETR can be easily integrated into other existing DETR variants without introducing extra parameters or computation and result in a significant improvement, particularly for small and large objects, demonstrating its generalization and effectiveness.

\vfill\pagebreak

\bibliographystyle{IEEEbib}
\bibliography{strings,refs}

\end{document}